%% file: 1-paper.tex
\documentclass{article} 
\usepackage{iclr2024_conference,times}

\input{inc-math}

\input{inc-packages}
\input{inc-macros}

\usepackage{hyperref}
\usepackage{url}

\title{Learning adaptive planning representations with natural language guidance}

\author{\textbf{Lionel Wong}\normalfont{\textsuperscript{1}\thanks{Asterisk indicates equal contribution. Correspondence to \texttt{zyzzyva@mit.edu}. Code for this paper will be released upon publication.}
}\quad
\textbf{Jiayuan Mao}\textsuperscript{1*}\quad 
\textbf{Pratyusha Sharma}\textsuperscript{1*}\quad 
\textbf{Zachary S. Siegel}\textsuperscript{2}\quad
\textbf{Jiahai Feng}\textsuperscript{3} \\
\textbf{Noa Korneev}\textsuperscript{4}\quad 
\textbf{Joshua B. Tenenbaum}\textsuperscript{1}\quad
\textbf{Jacob Andreas}\textsuperscript{1}
 \\
\textsuperscript{1}MIT\quad\textsuperscript{2}Princeton University \quad \textsuperscript{3}UC Berkeley \quad \textsuperscript{4}Microsoft\\
}

\iclrfinalcopy 
\begin{document}

\maketitle

\input{text/0-abstract}
\input{text/1-intro}

\input{text/3-method}
\input{text/4-experiment}

\input{text/2-related}
\input{text/5-conclusion}

\bibliography{ref}
\bibliographystyle{iclr2024_conference}

\clearpage

\appendix
\section{Appendix}
We will release a complete code repository containing our full algorithm implementation, all baselines, and benchmark tasks. Here, we provide additional details on our implementational choices.

\subsection{Additional Methods Implementation Details}
\subsubsection{LLM Prompting}\label{sec:appendix-llm-prompting}
We use \texttt{gpt-3.5-turbo-16k} for all experiments and baselines. Here, we describe the contents of the LLM few-shot prompts used in our method in more detail. 
\xhdr{Symbolic Task Decomposition}
For all unsolved tasks, at each iteration, we sample a set of symbolic task descriptions as a sequence of named high-level actions and their arguments.
We construct a few-shot prompt consisting of the following components:
\begin{enumerate}
    \item A brief natural language header (\textit{;;;; Given natural language goals, predict a sequence of PDDL actions});
    \item A sequence of example ($l_t$, $P_A$) tuples containing linguistic goals and example task decompositions. To avoid biasing the language model in advance, we provide example task decompositions for similar, constructed tasks that do not use any of the skills that need to be learned in our two domains. 
    
    For example, on ALFRED, these example task decompositions are for example tasks (\textit{bake a potato and put it in the fridge}, \textit{place a baked, grated apple on top of the dining table}, \textit{ place a plate in a full sink.}, and \textit{pick up a laptop and then carry it over to the desk lamp, then restart the desk lamp.}), and our example task decompositions suggest named operators \textit{BakeObject}, \textit{GrateObject}, \textit{FillObject}, and \textit{RestartObject}, none of which appear in the actual training set.
    \item At iterations $>0$, we also provide a sequence of sampled ($l_t$, $P_A$) tuples randomly sampled from any solved tasks and their discovered high-level plans. This means that few-shot prompting better represents the true task distribution over successive iterations.
\end{enumerate}
In our experiments, we prompt with temperature=1.0 and draw n=4 task decomposition samples per unsolved task.

\xhdr{Symbolic Operator Definition}
For all unsolved tasks, at each iteration, we sample proposed operator definitions consisting of $\textit{args}, \textit{pre}, \textit{eff}$ conditioned on all undefined operator names that appear in the proposed task decompositions.

For each operator name, we construct a few-shot prompt consisting of the following components:
\begin{enumerate}
    \item A brief natural language header (\textit{You are a software engineer who will be writing planning operators in the PDDL planning language. These operators are based on the following PDDL domain definition.}
    \item The full set of environment predicates vocabulary of high-level environment predicates $\gP$, as well as valid named argument values (eg. object types).
    \item A sequence of example $\textit{name}, \textit{args}, \textit{pre}, \textit{eff}$ operator definitions demonstrating the PDDL definition format. As with task decomposition, of course, we do not provide any example operator definitions that we wish to learn from our dataset.
    
    \item At iterations $>0$, we include as many possible validated $\textit{name}, \textit{args}, \textit{pre}, \textit{eff}$ operators defined in the current library (including new learned operators). If there are shared patterns between operators, this means that few-shot prompting also better represents the true operator structure over successive iterations.
\end{enumerate}
In our experiments, we prompt with temperature=1.0 and draw n=3 task decomposition samples per unsolved task.
However, in our pilot experiments, we actually find that sampling directly from the token probabilities defined by this few-shot prompt does not produce sufficiently diverse definitions for each operator name. We instead directly prompt the LLM to produce up to N distinct operator definitions sequentially.

We find that GPT 3.5 frequently produces syntactically invalid operator proposals -- proposed operators often include invent predicates and object types that are not defined in the environment vocabulary, do not obey the predicate typing rules, or do not have the correct number and types of arguments. While this might improve with finetuned or larger LLMs, we instead implement a simple post-processing heuristic to correct operators with syntactic errors, or reject operators altogether: as operator pre and postconditions are represented as conjunctions of predicates, we remove any invalid predicates (predicates that are invented or that specify invalid arguments); we collect all arguments named across the predicates and use the ground truth typing to produce the final $\textit{args}$, and we reject any operators that have 0 valid postcondition predicates. This post-processing procedure frequently leaves operators underspecified (\eg, the resulting operators now are missing necessary preconditions, which were partially generated but syntactically incorrect in the proposal); we allow our full operator learning algorithm to verify and reject these operators.

\xhdr{Symbolic Goal Proposal}
Finally, as described in \ref{sec:goal-proposal}, we also use an LLM to propose a set of candidate goal definitions as FOL formulas $F_t'$ defined over the environment predicates $\gP$ for each task. Our prompting technique is very similar to that used in the rest of our algorithm. For each task, we we construct a few-shot prompt consisting of the following components:
\begin{enumerate}
    \item A brief natural language header (\textit{You are a software engineer who will be writing goal definitions for a robot in the PDDL planning language.}
    \item The full set of environment predicates vocabulary of high-level environment predicates $\gP$, as well as valid named argument values (eg. object types).
    \item A sequence of example $l_t, f_t$ language and FOL goal formulas. In our experiments, during training, unlike in the previous prompts (where including ground truth operators would solve the learning problem), we \textit{do} sample an initial set of goal definitions from the training distribution as our initial example supervision. We set supervision to a randomly sampled fraction (0.1) of the training distribution. 
    
    \item At iterations $>0$, we also include $l_t, f_t$ examples from successfully solved tasks.
\end{enumerate}
In our experiments, we prompt with temperature=1.0 and draw n=4 task decomposition samples per unsolved task.
As with the operator proposal, we also find that sampling directly from the token probabilities defined by this few-shot prompt does not produce sufficiently diverse definitions for each linguistic goal to correct for ambiguity in the human language (eg. to define the multiple concrete \textit{Table} types that a person might mean when referring to a \textit{table}). We therefore again instead directly prompt the LLM to produce up to N distinct operator definitions sequentially.

We also post-process proposed goals using the same syntactic criterion to remove invalid predicates in the FOL formula, and reject any empty goals.

\input{fig-text/gpt4-tab-minecraft-alfred-combined}

\subsubsection{Policy Learning and Guided Low-level Search}
Concretely, we implement our policy-guided low-level action search as the following. We maintain a dictionary $D$ that maps subgoals (a conjunction of atoms) to a set of candidate low-level action trajectories. When planning for a new subgoal $sg$, if $D$ contains the trajectory, we prioritize trying candidate low-level trajectories in $D$. Otherwise, we fall back to a brute-force breadth-first search over all possible action trajectories. To populate $D$, during the BFS, we compute the difference in the environment state before and after the agent executes any sampled trajectory and the corresponding trajectory $t$ that caused the state change. Here the state difference can be viewed as a subgoal $sg$ achieved by executing $t$. Rather than directly adding the $(sg, t)$ as a key-value pair to $D$, we \textit{lift} the trajectory and environment state change by replacing concrete objects in $sg$ and $t$ by variables. Note that we update $D$ with each sampled trajectory in the BFS even if it doesn't achieve the subgoal specified in the BFS search.

When the low-level search receives a subgoal $sg$, we again lift it by replacing objects with variables, and try to match it with entries in $D$. If $D$ contains multiple trajectories $t$ for a given subgoal $sg$, we track how often a given trajectory succeeds for a subgoal and prioritize trajectories with the most successes.



\subsection{Experiments}\label{sec:appendix-experiments}

\xhdr{Learned Operator Libraries on Minecraft}
The following shows the full PDDL domain definition including the initial provided vocabulary of symbolic environment constants and predicates, initial pick and place operators and example operator, and all ensuing learned operators combined from the \textbf{Mining} and \textbf{Crafting} benchmarks.
\lstinputlisting[breaklines,language=lisp]{figs/\_cw\_final\_generated.pddl}

\xhdr{Learned Operator Libraries on ALFRED}
The following shows the full PDDL domain definition including the initial provided vocabulary of symbolic environment constants and predicates, initial pick and place operators, and all ensuing learned operators.
\lstinputlisting[breaklines,language=lisp]{figs/alfred\_linearized\_learned\_operators.pddl}

\end{document}

%% file: inc-math.tex
\usepackage{amsmath,amsfonts,bm}

\def\eqref#1{equation~\ref{#1}}

\def\1{\bm{1}}

\DeclareMathAlphabet{\mathsfit}{\encodingdefault}{\sfdefault}{m}{sl}
\SetMathAlphabet{\mathsfit}{bold}{\encodingdefault}{\sfdefault}{bx}{n}

\def\gA{{\mathcal{A}}}

\def\gP{{\mathcal{P}}}

\def\gS{{\mathcal{S}}}
\def\gT{{\mathcal{T}}}
\def\gU{{\mathcal{U}}}

\def\gX{{\mathcal{X}}}



%% file: inc-packages.tex
\usepackage{color,xcolor}
\usepackage{epsfig}
\usepackage{graphicx}

\usepackage{adjustbox}
\usepackage{array}
\usepackage{booktabs}
\usepackage{colortbl}
\usepackage{wrapfig}
\usepackage{hhline}
\usepackage{multirow}
\usepackage{subcaption} 
\usepackage{wrapfig}
\usepackage{floatflt}

\usepackage{amsmath,amsfonts,amssymb,amsthm}
\usepackage{mathtools}  
\usepackage{bm}
\usepackage{nicefrac}
\usepackage{microtype}

\usepackage{changepage}
\usepackage{extramarks}
\usepackage{fancyhdr}
\usepackage{lastpage}
\usepackage{setspace}
\usepackage{soul}
\usepackage{xspace}
\usepackage[font=small]{caption}

\usepackage[pagebackref=true,breaklinks=true,colorlinks,citecolor=gray]{hyperref}

\usepackage{url}

\usepackage{enumerate}
\usepackage{todonotes} 
\usepackage{enumitem}  

\usepackage{titlesec}

\usepackage{makecell}

\usepackage{pifont} 

\usepackage{algorithm}
\usepackage[noend]{algpseudocode}

\usepackage{listings}
\usepackage{framed}
\definecolor{shadecolor}{gray}{0.95}

\usepackage{xparse}
\usepackage{etoolbox}

%% file: inc-macros.tex
\newcolumntype{L}[1]{>{\raggedright\let\newline\\\arraybackslash\hspace{0pt}}m{#1}}
\newcolumntype{C}[1]{>{\centering\let\newline\\\arraybackslash\hspace{0pt}}m{#1}}
\newcolumntype{R}[1]{>{\raggedleft\let\newline\\\arraybackslash\hspace{0pt}}m{#1}}

\newcommand{\sect}[1]{Section~\ref{#1}}

\newcommand{\fig}[1]{Fig.~\ref{#1}}
\newcommand{\tbl}[1]{Table~\ref{#1}}

\newcommand{\ignore}[1]{}

\newcommand{\cmark}{\ding{51}}%
\newcommand{\xmark}{\ding{55}}%

\makeatletter
\DeclareRobustCommand\onedot{\futurelet\@let@token\@onedot}
\def\@onedot{\ifx\@let@token.\else.\null\fi\xspace}

\def\eg{e.g\onedot} 
\def\ie{i.e\onedot} 
 
\def\etc{etc\onedot}

\makeatother

\definecolor{MyDarkBlue}{rgb}{0,0.08,1}
\definecolor{MyDarkGreen}{rgb}{0.02,0.6,0.02}
\definecolor{MyDarkRed}{rgb}{0.8,0.02,0.02}
\definecolor{MyDarkOrange}{rgb}{0.40,0.2,0.02}
\definecolor{MyPurple}{RGB}{111,0,255}
\definecolor{MyRed}{rgb}{1.0,0.0,0.0}
\definecolor{MyGold}{rgb}{0.75,0.6,0.12}
\definecolor{MyDarkgray}{rgb}{0.66, 0.66, 0.66}
\definecolor{JiayuanColor}{rgb}{0.60,0.43,0.48}

\newcommand{\model}{Ada\xspace}

\newcommand{\object}[1]{\textit{#1}}

\NewDocumentCommand{\prop}{m >{\SplitList{,}}m}{%
  \ensuremath{\textit{#1}(%
  \ProcessList{#2}{\propositionitem}%
  )}%
  \propositionfirstitemtrue
}

\newif\ifpropositionfirstitem
\propositionfirstitemtrue  
\newcommand\propositionitem[1]{%
  \ifpropositionfirstitem
    \propositionfirstitemfalse
  \else
    , %
  \fi
  \text{#1}}

\newcommand{\xhdr}[1]{\noindent\textbf{#1}}

\theoremstyle{plain}



%% file: text/0-abstract.tex
\begin{abstract}
Effective planning in the real world requires not only world knowledge, but the ability to leverage that knowledge to build the \textit{right representation} of the task at hand. Decades of hierarchical planning techniques have used domain-specific temporal \textit{action abstractions} to support efficient and accurate planning, almost always relying on human priors and domain knowledge to decompose hard tasks into smaller subproblems appropriate for a goal or set of goals. 
This paper describes \textit{Ada} (Action Domain Acquisition), a framework for automatically constructing task-specific planning representations using task-general background knowledge from language models (LMs). Starting with a general-purpose hierarchical planner and a low-level goal-conditioned policy, Ada
\textbf{interactively learns a library of planner-compatible high-level action abstractions and low-level controllers adapted to a particular domain of planning tasks}. 
On two language-guided interactive planning benchmarks (\textit{Mini Minecraft} and \textit{ALFRED Household Tasks}), Ada strongly outperforms other approaches that use LMs for sequential decision-making, offering more accurate plans and better generalization to complex tasks.

\end{abstract}

%% file: text/1-intro.tex
\vspace{-0.5em}
\section{Introduction}
\vspace{-0.5em}

People make complex plans over long timescales, flexibly adapting what we \textit{know} about the world in general to govern how we act in specific situations. To make breakfast in the morning, we might convert a broad knowledge of cooking and kitchens into tens of fine-grained motor actions in order to find, crack, and fry a specific egg; to achieve a complex research objective, we might plan a routine over days or weeks that begins with the low-level actions necessary to ride the subway to work. The problem of \textit{adapting general world knowledge to support flexible long-term planning} is one of the unifying challenges of AI. While decades of research have developed representations and algorithms for solving restricted and shorter-term planning problems, generalized and long-horizon planning remains a core, outstanding challenge for essentially all AI paradigms, including classical symbolic planning~\citep{erol1994htn}, reinforcement learning~\citep{sutton1999between}, and modern generative AI~\citep{wang2023voyager}.

How do humans solve this computational challenge? A growing body of work in cognitive science suggests that people come up with \textit{hierarchical, problem-specific representations} of their actions and environment to suit their goals, tailoring how they represent, remember, and reason about the world to plan efficiently for a particular set of tasks ~\citep[\eg,][]{ho2022people}. In AI, a large body of work has studied \textit{hierarchical planning using domain-specific temporal abstractions}---progressively decomposing high-level goals into sequences abstract actions that eventually bottom out in low-level control. An extensive body of work has explored how to plan using these hierarchical action spaces, including robotic task-and-motion planning (TAMP) systems~\citep{garrett2021integrated} and hierarchical RL frameworks~\citep{sutton1999between}. 

However, identifying a set of abstract actions that are relevant and useful for achieving any given set of goals remains the central bottleneck in general hierarchical planning. Intuitively, ``useful" high-level actions must satisfy many different criteria: they should enable time-efficient high-level planning, correspond feasible low-level action sequences, and compose and generalize to new tasks. Despite efforts to learn high-level actions automatically in both classical planning~\citep{nejati2006learning} and RL formulations~\citep{dietterich2000hierarchical}, most state-of-the-art robotics and planning systems rely on human expertise to hand-engineer new planning representations for each new domain \citep{ahn2022can}. 

\input{fig-text/teaser}
 
In this paper, we introduce {\it Action Domain Acquisition} (\textit{Ada}), a framework for using background knowledge from language (conveyed via language models) as an initial source of task-relevant domain knowledge. Ada uses language models (LMs) in an interactive planning loop to assemble a \textbf{library of composable, hierarchical actions 
tailored to a given environment and task space}. 
Each action consists of two components: (1) a \textit{high-level abstraction} represented as a symbolic planning \textit{operator}~\citep{fikes1971strips} that specifies preconditions and action effects as sets of predicates; and (2) a \textit{low-level controller} that can achieve the action's effects by predicting a sequence of low-level actions with a neural network or local search procedure.
We study planning in a multitask reinforcement learning framework, in which agents interact with their environments to must solve collections of tasks of varying complexity. 
Through interaction, Ada builds up a library of actions incrementally, at each step ensuring that learned high-level actions compose to produce valid abstract plans and realizable low-level trajectories.

We evaluate Ada (\fig{fig:splash}) on two benchmarks, \textit{Mini Minecraft} and \textit{ALFRED} ~\citep{shridhar2020alfred}. We compare this approach against three baselines that leverage LMs for sequential decision-making in other ways: to parse linguistic goals into formal specifications that are solved directly by a planner (as in ~\cite{liu2023llmp}), to directly predict sequences of high-level subgoals (as in ~\cite{ahn2022can}), and to predict libraries of actions defined in general imperative code (as in ~\cite{wang2023voyager}). In both domains, we show that Ada learns action abstractions that allow it to solve dramatically more tasks on each benchmark than these baselines, and that these abstractions compose to enable efficient and accurate planning in complex, unseen tasks.

%% file: fig-text/teaser.tex
\begin{figure}[tp]
\begin{center}
\includegraphics[width=1.0\textwidth]{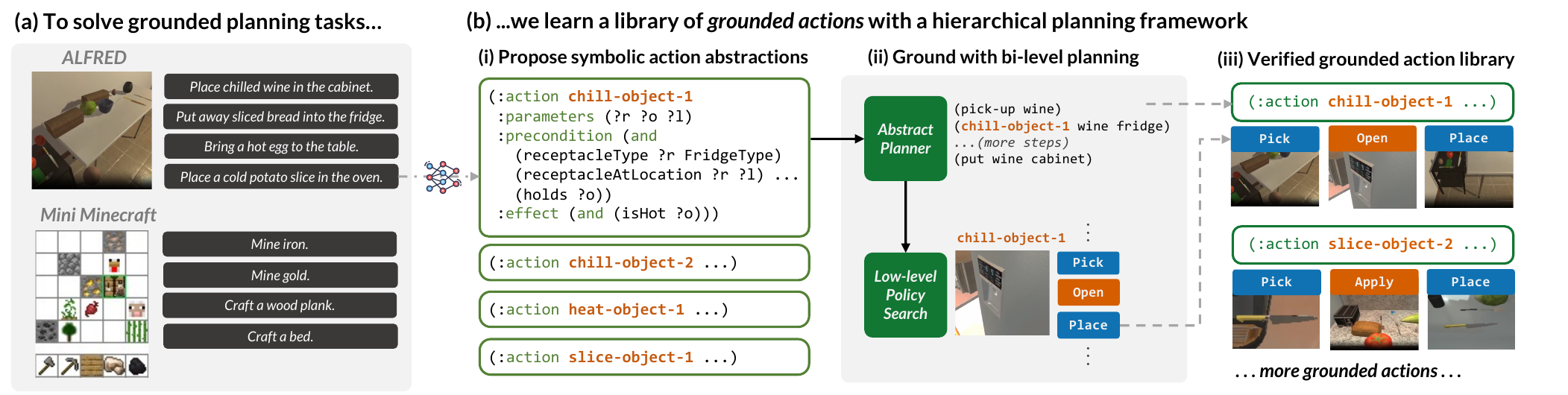}
\end{center}
\vspace{-1em}
\caption{Our approach solves complex planning tasks specified in language and grounded in interactive environments by jointly learning a \textit{library of symbolic high-level action abstractions and modular low-level controllers} associated with each abstraction. Our system leverages background information in language as a prior to \textit{propose useful action abstractions}, then uses a \textit{hierarchical planning framework} to verify and ground them.}
\label{fig:splash}
\vspace{-1em}
\end{figure}

%% file: text/3-method.tex
\section{Problem Formulation}

\label{sec:problem-formulation}
We assume access to an environment $\langle \gX, \gU, \gT \rangle$, where $\gX$ is the (raw) state space, $\gU$ is the (low-level) action space (\eg, robot commands), and $\gT$ is a deterministic transition function $\gT: \gX \times \gU \rightarrow \gX$. We also have a set of features (or ``predicates'') $\gP$ that define an abstract state space $\gS$: each abstract state $s \in \gS$ is composed of a set of objects and their features. For example, a simple scene that contains a red box on a table could be represented by an abstract state containing two objects $\object{A}$ and $\object{B}$, and a set of atoms $\{\prop{red}{A}, \prop{box}{A}, \prop{table}{B},\prop{on}{A,B}\}$. We assume the mapping from environmental states to abstract states $\Phi: \gX \rightarrow \gS$ is given and fixed (though see \citealp{migimatsu2022grounding} for how it might be learned).

\input{fig-text/representation}
\input{fig-text/framework}

In addition to the environment, we have a collection of tasks $t$. Each $t$ is described by a natural language instruction $\ell_t$, corresponding to a goal predicate (which is not directly observed). In this paper, we assume that predicates may be defined in terms of abstract states, \ie, $g_t: \gS \rightarrow \{T, F\}$.
Our goal is to build an agent that, given the initial state $x_0 \in \gX$ and the natural language instruction $\ell_t$, can generate a sequence of low-level actions $\{u_1, u_2, \cdots, u_H\} \in \gU^H$ such that $g_t(\Phi(x_H))$ is true (where $x_H$ is the terminal state of sequentially applying $\{u_i\}$ on $x_0$). The agent receives reward signal only upon achieving the goal specified by $g_t$.

Given a very large number of interactions, a sufficiently expressive reflex policy could, in principle, learn a policy that maps from low-level states to low-level actions conditioned on the language instruction $\pi(u \mid x; \ell_t)$. However, for very long horizons $H$ and large state spaces (\eg, composed of many objects and compositional goals), such algorithms can be highly inefficient or effectively infeasible. 
The key idea behind our approach is to use
natural language descriptions $\ell_t$ to bootstrap a high-level action space $\gA$ over the abstract state space $\gS$ to accelerate learning and planning.

Formally, our approach learns a library of high-level actions (also called \textit{operators} in classical planning) $\gA$. As illustrated in \fig{fig:representation}b, each $a \in \gA$ is a tuple of $\langle \textit{name}, \textit{args}, \textit{pre}, \textit{eff}, \textit{controller} \rangle$. $\textit{name}$ is the name of the action, $\textit{args}$ is a list of variables, usually denoted by $?x, ?y, \etc$, $\textit{pre}$ is a precondition formula based on the variables $\textit{args}$ and the features $\gP$, and $\textit{eff}$ is the effect, which is also defined in terms of $\textit{args}$ and $\gP$. Finally, $\textit{controller}: \gX \rightarrow \gU$ is a low-level policy associated with the action. The semantics of the preconditions and effects is: for any state $x$ such that $\textit{pre}(\Phi(x))$, executing $\textit{controller}$ starting in $x$ (for an indefinite number of steps) will yield a state $x'$ such that $\textit{eff}(\Phi(x'))$~\citep{lifschitz1987semantics}. \textbf{In this framework, $\gA$ defines a partial, abstract world model of the underlying state space.}

As shown in \fig{fig:representation}b, given the set of high-level actions and a parse of the instruction $\ell_t$ into a first-order logic formula, we can leverage symbolic planners \citep[\eg,][]{helmert2006fast} to first compute a high-level plan $\{a_1, \cdots, a_K\} \in \gA^K$ that achieves the goal $\ell_t$ symbolically, and then refine the high-level plan into a low-level plan leveraging the controllers associated with each action. This bi-level planning approach decomposes long-horizon planning problems into several short-horizon problems. Furthermore, it can also leverage the compositionality of high-level actions $\gA$ to generalize to longer plans.

\section{Action Abstractions from Language}

As illustrated in \fig{fig:framework}, our framework, {\it Action Domain Acquisition} ({\it \model}) learns action abstractions iteratively as it attempts to solve tasks. Our algorithm is given a dataset of tasks and their corresponding language descriptions, the feature set $\gP$, and optionally an initial set of high-level action operators $\gA_0$. At each iteration $i$, we first use a large language model (LLM) to propose a set of novel high-level action definitions $\gA_i'$ based on the features $\gP$ and the language goals $\{\ell_t\}$ (\sect{sec:operator-proposal}). Next, we use a LLM to also translate each language instruction $\ell_t$ into a symbolic goal description $F_t$, and use a bi-level planner to compute a low-level plan to accomplish $\ell_t$ (\sect{sec:bilevel}). Then, based on the planning and execution results, we score each operator in $\gA_i$ and add ones to the verified library if they have yielded successful execution results (\sect{sec:score}). To accelerate low-level planning, we simultaneously learn local subgoal-conditioned policies (\ie, the controllers for each operator; \sect{sec:policy-learning}). Algorithm~\ref{alg:overall} summarizes the overall framework.

A core goal of our approach is to adapt the initial action abstractions proposed from an LLM prior into a set of \textit{useful} operators $\gA*$ that permit efficient and accurate planning on a dataset of tasks and ideally, that generalize to future tasks. While language provides a key initial prior, our formulation refines and verifies the operator library to adapt to a given planning procedure and environment (similar to other action-learning formulations like ~\citealp{silver2021learning}). Our formulation ensures not only that the learned operators respect the dynamics of the environment, but also fit their grain of abstraction according to the capacity of the controller, trading off between fast high-level planning and efficient low-level control conditioned on each abstraction.

\input{fig-text/framework-algo}
\input{fig-text/prompt}

\subsection{Operator Proposal}
\label{sec:operator-proposal}
\begin{center}
\vspace{-1.5em}
\[ \gA_i \leftarrow \gA_{i-1} \cup \text{ProposeOperatorDefinitions}(\gP, \{\ell_t\})\]
\end{center}
At each iteration $i$, we use a pretrained LLM to extend the previous operator library $\gA_{i-1}$ with a large set of candidate operator definitions proposed by the LLM based on the task language descriptions and environment features $\gP$. This yields an extended candidate library $\gA_i'$ where each $a \in \gA_i' = \langle \textit{name}, \textit{args}, \textit{pre}, \textit{eff} \rangle$ where \textit{name} is a human-readable action name and $\textit{args}, \textit{pre}, \textit{eff}$ are a PDDL operator definition. We employ a two-stage prompting strategy: symbolic task decomposition followed by symbolic operator definition.

\xhdr{Example.} \fig{fig:prompt} shows a concrete example. Given a task instruction (\textit{Bring a hot egg to the table}) and the abstract state description, we first prompt the LLM to generate an abstract task decomposition, which may contain operator names that are undefined in the current operator library. Next, we extract the names of those undefined operators and prompt LLMs to generate the actual symbolic operator descriptions, in this case, the new \textit{heat-object} operator.

\xhdr{Symbolic task decomposition.} For a given task $\ell_t$ and a initial state $x_0$, we first translate the raw state $x_0$ into a symbolic description $\Phi(x_0)$. To constrain the length of the state description, we only include unary features in the abstract state (\ie, only object categories and properties). Subsequently, we present a few-shot prompt to the LLM and query it to generate a proposed task decomposition conditioned on the language description $\ell_t$. It generates a sequence of named high-level actions and their arguments, which explicitly can include high-level actions that are not yet defined in the current action library. We then extract all the operator names proposed across tasks as the candidate high-level operators. Note that while in principle we might use the LLM-proposed task decomposition itself as a high-level plan, we find empirically that this is less accurate and efficient than a formal planner.

\xhdr{Symbolic operator definition.} With the proposed operator names and their usage examples (\ie, the actions and their arguments in the proposed plans), we then few-shot prompt the LLM to generate candidate operator \textit{definitions} in the PDDL format (argument types, and pre/postconditions defined based on features in $\gP$). We also post-process the generated operator definitions to remove feature names not present in $\gP$ and correct syntactic errors. We describe implementation details for our syntax correction strategy in the appendix.

\subsection{Goal Proposal and Bi-Level Planning}
\label{sec:goal-proposal}
\label{sec:bilevel}
\begin{center}
\vspace{-1.5em}
\begin{align*} &\textbf{for}~\text{each unsolved task $(x^{(j)}_0, \ell^{(j)}_t)$:} \\[-.4em]
&\qquad \bar{u} \leftarrow \text{BiLevelPlan}(\gA_i, \ell^{(j)}_t, \pi) \\[-.4em]
&\qquad \textit{result}^{(j)} \leftarrow \text{Execute}(x^{(j)}_0, \bar{u})
\end{align*}
\end{center}
At each iteration $i$, we then attempt to \textit{BiLevelPlan} for unsolved tasks in the dataset. This step attempts to find and execute a low-level action sequence $\{u_1, u_2, \cdots, u_H\} \in \gU^H$ for each task using the proposed operators in $\gA_i'$ that satisfies the unknown goal predicate $g_t$ for each task. This provides the environment reward signal for action learning. Our \textit{BiLevelPlan} implementation uses standard hierarchical planning tools as follows:

\xhdr{Symbolic goal proposal:} As defined in Sec. \ref{sec:problem-formulation}, each task is associated with a queryable but unknown goal predicate $g_t$ that can be represented as a first-order logic formula $f_t$ over symbolic features in $\gP$. Our agent only has access to a linguistic task description $\ell_t$, so we use a few-shot prompted LLM to predict candidate goal formulas $F_t'$ conditioned on $\ell_t$ and features $\gP$.

\xhdr{High-level planning}: Given each candidate goal formula $f_t' \in F_t'$, the initial abstract problem state $s_0$, and the current candidate operator library $\gA'$, we search for a \textit{high-level plan} $P_{A} = \{(a_1, o_{1_i}...), \cdots, (a_K, o_{K_i}...)\}$ as a sequence of high-level actions from $\gA'$ concretized with object arguments $o$, such that executing the action sequence would satisfy $f_t'$ according to the operator definitions. This is a standard symbolic PDDL planning formulation; we use an off-the-shelf symbolic planner, FastDownward~\citep{helmert2006fast} to find high-level plans.

\xhdr{Low-level planning and environment feedback}: We then search for a low-level plan as a sequence of low-level actions $\{u_1, u_2, \cdots, u_H\} \in \gU^H$, conditioned on the high-level plan structure. Each concretized action tuple $(a_i, o_{1_i}...) \in P_{A}$ defines a local subgoal $sg_i$, as the operator postcondition parameterized by the object arguments $o$. For each $(a_i, o_{1_i}...) \in P_{A}$, we therefore search for a sequence of low-level actions $u_{i_1}, u_{i_2}...$ that satisfies the local subgoal $sg_i$. We search with a fixed budget per subgoal, and fail early if we are unable to satisfy the local subgoal $sg_i$. If we successfully find a complete sequence of low-level actions satisfying all local subgoals $sg_i$ in $P_{A}$, we execute all low-level actions and query the hidden goal predicate $g_t$ to determine environment reward. We implement a basic learning procedure to simultaneously learn subgoal-conditioned controllers over time (described in \sect{sec:policy-learning}), but our formulation is general and supports many hierarchical planning schemes (such as sampling-based low-level planners~\citep{lavalle1998rapidly} or RL algorithms).

\subsection{Policy Learning and Guided Low-Level Search}
\label{sec:policy-learning}
\begin{center}
\vspace{-1.5em}
\[\theta \leftarrow \text{UpdateSubgoalPolicy}(\theta, \textit{result})\]
\end{center}
The sequence of subgoals $sg_i$ corresponding to high-level plans $P_{A}$ already restricts the local low-level planning horizon. However, we further learn subgoal-conditioned low-level policies $\pi(u | x; \textit{sg})$ from environment feedback during training to accelerate low-level planning. To exploit shared structure across subgoals, we learn a shared controller for all operators from $x \in \gX$ and conjunctions of predicates in $sg$. To maximize learning during training, we use a hindsight goal relabeling scheme~\citep{andrychowicz2017hindsight}, supervising on all conjunctions of predicates in the state as we roll out low-level search. While the shared controller could be learned as a supervised neural policy, we find that our learned operators sufficiently restrict the search to permit learning an even simpler count-based model from $X, sg \rightarrow u \in \gU$. We provide additional details in the Appendix.

\subsection{Scoring LLM Operator Proposals}
\label{sec:score}
\begin{center}
\vspace{-1.5em}
\[\gA_i \leftarrow \text{ScoreAndFilter}(\gA_i, \textit{result})\]
\end{center}
Finally, we update the learned operator library $\gA_{i}$ to retain candidate operators that were useful and successful in bi-level planning. Concretely, we estimate operator candidate $a_i' \in \gA_{i}'$ accuracy across the bi-level plan executions as $s/b$ where $b$ counts the total times $a_i'$ appeared in a high-level plan and $s$ counts successful execution of the corresponding low-level action sequence to achieve the subgoal associated with $a_i'$. We retain operators if $b > \tau_b$ and $s/b > \tau_r$, where $\tau_b, \tau_r$ are hyperparameters. Note that this scoring procedure learns whether operators are accurate and support low-level planning independently of whether the LLM-predicted goals $f_t'$ matched the true unknown goal predicates $g_t$.

%% file: fig-text/representation.tex
\begin{figure}[tp]
\small
\centering
\includegraphics[width=1.0\textwidth]{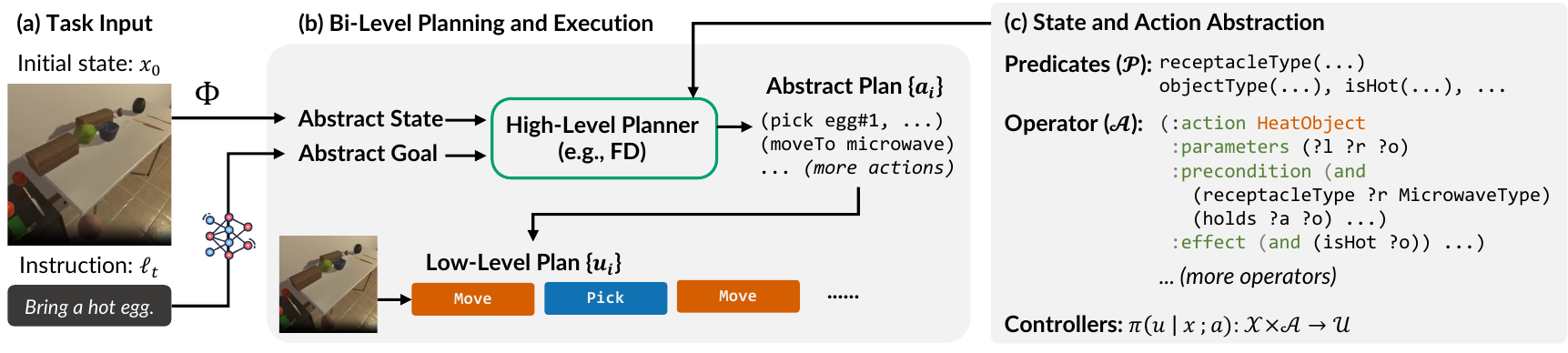}
\vspace{-1.5em}
\caption{Representation for our (a) task input, (b) the bi-level planning and execution pipeline for inference time, and (c) the abstract state and action representation.}
\label{fig:representation}
\end{figure}

%% file: fig-text/framework.tex
\begin{figure}[tp]
\small
\begin{center}
\includegraphics[width=1.0\textwidth]{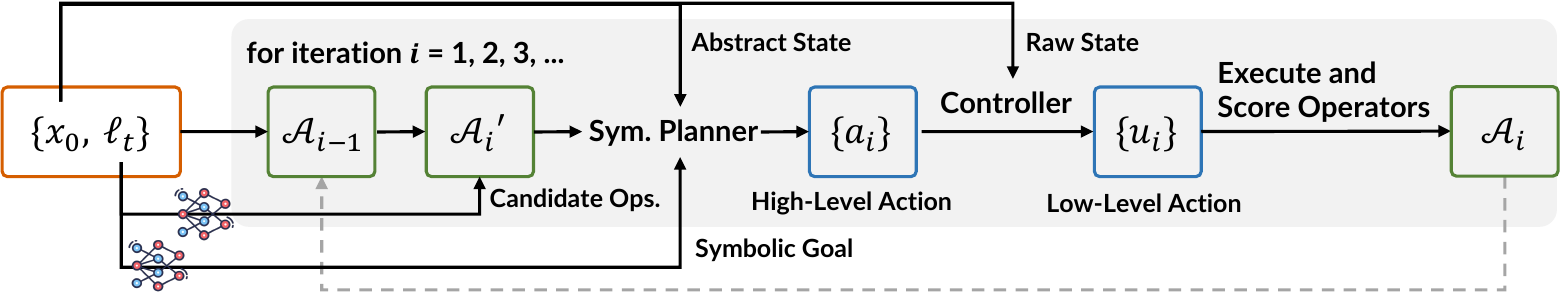}
\end{center}
\vspace{-1em}
\caption{The overall framework. Given task environment states and descriptions, at each iteration, we first propose candidate abstract actions (operators) $\gA_i'$, then uses bi-level planning and execution to solve tasks. We add operators to the operator library based on the execution result.}\vspace{-1.5em}
\label{fig:framework}

\end{figure}

%% file: fig-text/framework-algo.tex
\begin{algorithm}[tp]
\caption{Action Abstraction Learning from Language}\small
\label{alg:overall}
\begin{algorithmic}[1]
\Require Dataset of tasks and their language descriptions $\{\ell_t\}$
\Require Predicate set $\gP$
\Require Optionally, an initial set of abstract operators $\gA_0$, or $\gA_0 = \emptyset$
\State Initialize subgoal-conditioned policy $\pi_\theta$.
\For{$i = 1, 2, \cdots, M$}
    \State $\gA_i \leftarrow \gA_{i-1} \cup \text{ProposeOperatorDefinitions}(\gP, \{\ell_t\})$
    \Comment{\sect{sec:operator-proposal}}
    \For{each unsolved task $j$: $(x^{(j)}_0, \ell^{(j)}_t)$}
        \State $\bar{u} \leftarrow \text{BiLevelPlan}(\gA_i, \ell^{(j)}_t, \pi)$
        \Comment{\sect{sec:bilevel}}
        \State $\textit{result}^{(j)} \leftarrow \text{Execute}(x^{(j)}_0, \bar{u})$ \Comment{Execute the plan}
    \EndFor
    \State $\theta \leftarrow \text{UpdateSubgoalPolicy}(\theta, \textit{result})$
    \Comment{\sect{sec:policy-learning}}
    \State $\gA_i \leftarrow \text{ScoreAndFilter}(\gA_i, \textit{result})$
    \Comment{\sect{sec:score}}
\EndFor
\Return $\gA_M$
\end{algorithmic}
\end{algorithm}

%% file: fig-text/prompt.tex
\begin{figure}[tp]
\begin{center}
\includegraphics[width=1.0\textwidth]{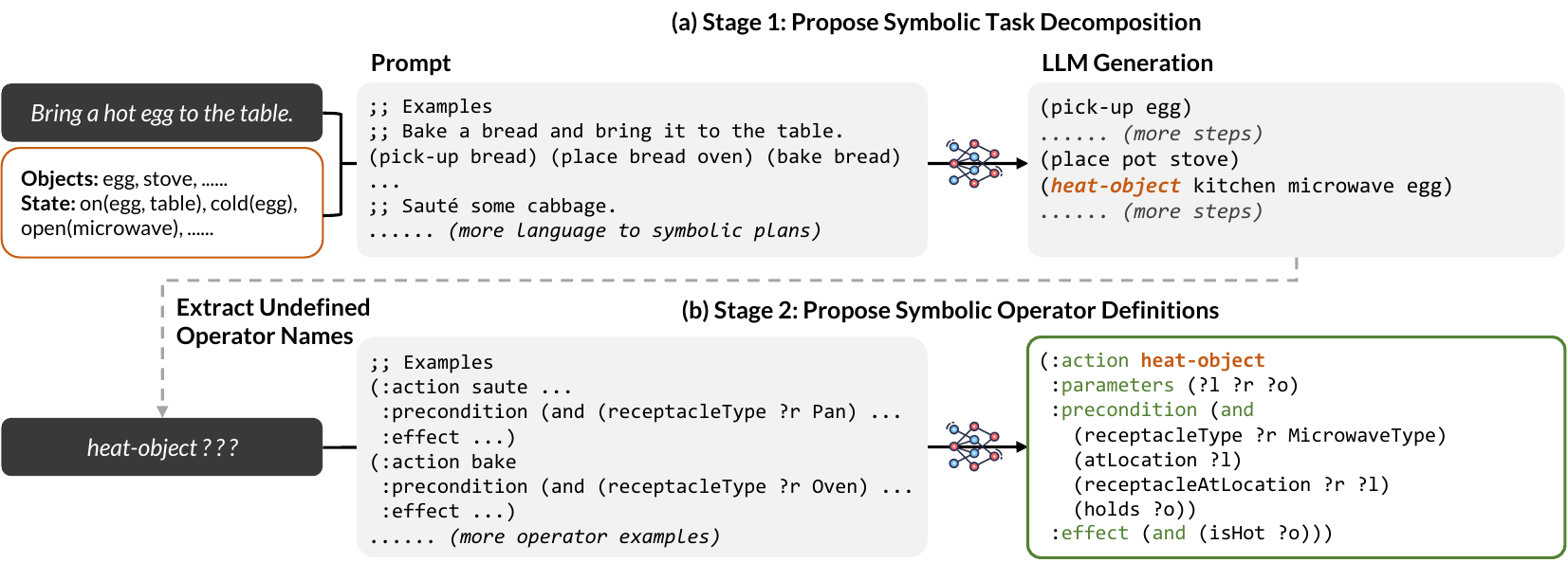}
\end{center}
\vspace{-1em}
\caption{Our two-stage prompting method for generating candidate operator definitions. (a) Given a task instruction, we first prompt an LLM to generate a candidate symbolic task decomposition. (b) We then extract undefined operator names that appear in these action sequences and prompt an LLM to generate symbolic definitions.}
\label{fig:prompt}
\vspace{-1.5em}
\end{figure}

%% file: text/4-experiment.tex
\vspace{-0.5em}
\section{Experiments}
\vspace{-0.5em}
\xhdr{Domains.} We evaluate our approach on two-language specified planning-benchmarks: \textit{Mini Minecraft} and \textit{ALFRED}~\citep{shridhar2020alfred}. 

\textit{Mini Minecraft} (\fig{fig:minecraft-alfred-combined}, \textit{top}) is a procedurally-generated Minecraft-like benchmark ~\citep{chen2020ask,luo2023learning} that requires complex, extended planning. The environment places an agent on a 2D map containing various resources, tools, and crafting stations. The agent can use appropriate tools to mine new items from raw resources (\eg use an \textit{axe} to obtain \textit{wood} from \textit{trees}), or collect resources into an inventory to craft new objects (\eg combining \textit{sticks} and \textit{iron ingots} to craft a \textit{sword}, which itself can be used to obtain \textit{feathers} from a \textit{chicken}). The ability to create new objects that themselves permit new actions yields an enormous action space at each time step ($>$2000 actions, considering different combinations of items to use) and very long-horizon tasks (26 steps for the most complex task, even without path-planning.) The provided environment predicates allow querying object types and inventory contents. Low-level actions allow the agent to move and apply tools to specific resources. To focus on complex crafting, we provide a low-level \textit{move-to} action to move directly to specified locations. Linguistic goal specifications are synthetically generated from a simple grammar over craftable objects and resources (\eg \textit{Craft a sword}, \textit{Mine iron ore}).

\textit{ALFRED} (\fig{fig:minecraft-alfred-combined}, \textit{bottom}) is a household planning benchmark of human-annotated but formally verifiable tasks defined over a simulated Unity environment \citep{shridhar2020alfred}. The interactive environment places an agent in varying 3D layouts, each containing appliances and dozens of household objects. The provided environment includes predicates for querying object types, object and agent locations, and classifiers over object states (eg. whether an object is \textit{hot} or \textit{on}). Low-level actions enable the agent to pick up and place objects, apply tools to other objects, and open, close, and turn on appliances. As specified in ~\cite{shridhar2020alfred}, ground-truth high-level plans in the ALFRED benchmark compose 5-10 high-level operators, and low-level action trajectories have on average 50 low-level actions. There over 100 objects that the agent can interact with in each interactive environment. 

As with Minecraft, we provide a low-level method to move the agent directly to specified locations. While ALFRED is typically used to evaluate detailed instruction following, we focus on a \textit{goal-only} setting that only uses the goal specifications. The human-annotated goals introduce ambiguity, underspecification, and errors with respect to the ground-truth verifiable tasks (eg. people refer to \textit{tables} without specifying if they mean the \textit{side table}, \textit{dining table}, or \textit{desk}; a \textit{light} when there are multiple distinct lamps; or a \textit{cabbage} when they want \textit{lettuce}). 

\input{fig-text/minecraft-alfred-combined}

\xhdr{Experimental setup.} We evaluate in an iterative continual learning setting; except on the compositional evaluations, we learn from \textit{n=2} iterations through all (randomly ordered) tasks and report final accuracy on those tasks. All experiments and baselines use GPT-3.5. For each task, at each iteration, we sample \textit{n=4} initial goal proposals and \textit{n=4} initial task decompositions, and \textit{n=3} operator definition proposals for each operator name. We report \textit{best-of} accuracy, scoring a task as solved if verification passes on at least one of the proposed goals. For Minecraft, we set the motion planning budget for each subgoal to $\leq$1000 nodes. For ALFRED, which requires a slow Unity simulation, we set it to 50 nodes. Additional temperature and sampling details are in the Appendix.

We evaluate on three \textit{Mini Minecraft} benchmark variations to test how our approach generalizes to complex, compositional goals. In the simplest \textbf{Mining} benchmark, all goals involve mining a target item from an appropriate initial resource with an appropriate tool (\eg, Mining \textit{iron} from \textit{iron\_ore} with an \textit{axe}), and require learning which tools can be used to obtain which items. In the harder \textbf{Crafting} benchmark, goals involve crafting a target artifact (\eg, a \textit{bed}), which may require mining a few target resources. The most challenging \textbf{Compositional} benchmark combines mining and crafting tasks, in environments that only begin with raw resources and two starting tools (axe and pickaxe). Agents may need to compose multiple skills to obtain other downstream resources (see \fig{fig:minecraft-alfred-combined} for an example). To test action generalization, we report evaluation on the \textit{Compositional} using \textit{only} actions learned previously in the \textbf{Mining} and \textbf{Crafting} benchmarks.

We similarly evaluate on an \textit{ALFRED} benchmark of \textbf{Simple and Compositional} tasks drawn from the original task distribution in \citet{shridhar2020alfred}. This distribution contains simple tasks that require picking up an object and placing it in a new location, picking up objects, applying a single household skill to an object and moving them to a new location (\eg, \textit{Put a clean apple on the dining table}), and compositional tasks that require multiple skills (\eg, \textit{Place a hot sliced potato on the counter}). We use a random subset of n=223 tasks, selected from an initial 250 that we manually filter to remove completely misspecified goals (which omit any mention of the target object or skill). 

\xhdr{Baselines.} We compare our method to three baselines, each related to language-guided planning systems in recent work. 

\textbf{\textit{Low-level Planning Only}} uses an LLM to predict only the symbolic goal specification conditioned on the high-level predicates and linguistic goal, then uses the low-level planner to search directly for actions that satisfy that goal. This baseline implements a model like \textbf{LLM+P} \citep{liu2023llmp}, which uses LLMs to translate linguistic goals into planning-compatible formal specifications, then attempt to plan directly towards these with no additional representation learning.

\textbf{\textit{Subgoal Prediction}} uses an LLM to predict a sequence of high-level subgoals (as PDDL pre/postconditions with object arguments), conditioned on the high-level predicates, and task goal and initial environment state. This baseline implements a model like \textbf{SayCan} ~\citep{ahn2022can}, which uses LLMs to directly predict goal \textit{and} a sequence of decomposed formal subgoal representations, then applies low-level planning over these formal subgoals. 

\textbf{\textit{Code Policy Prediction}} uses an LLM to predict the definitions of a library of \textit{imperative local code policies} in Python (with cases and control flow) over an imperative API that can query state and execute low-level actions.) Then, as FastDownward planning is no longer applicable, we also use the LLM to predict the function call sequences with arguments for each task. This baseline implements a model like \textbf{Voyager} \citep{wang2023voyager}, which uses an LLM to predict a library of skills implemented as imperative code and to directly predict a sequence of these skills to solve individual tasks. Like Voyager, we verify the individual code skills during interactive planning, but do not use a more global learning objective  to attempt to learn a concise or non-redundant library.

\vspace{-0.5em}
\subsection{Results}
\vspace{-0.5em}
\input{fig-text/tab-minecraft-alfred-combined}
\xhdr{What action libraries do we learn?}
\fig{fig:minecraft-alfred-combined} shows example operators learned on each domain (Appendix \ref{sec:appendix-experiments} contains the full libraries of operators learned on both domains from a randomly sampled run of the n=3 replications). In \textit{Mini Minecraft}, we manually inspect the library and find that we learn operators that correctly specify the appropriate tools, resources, and outputs for all intermediate mining actions (on \textbf{Mining}) and crafting actions (on \textbf{Crafting}), allowing perfect direct generalization to the \textbf{Compositional} tasks without any additional training on these complex tasks. In \textit{ALFRED}, we compare the learned libraries from all runs to the ground-truth operator library hand-engineered in ~\cite{shridhar2020alfred}. The ground-truth operator set contains 8 distinct operators corresponding to different compositional skills (eg. \textit{Slicing}, \textit{Heating}, \textit{Cleaning}, \textit{Cooling}). Across all replications, model reliably recovers semantically identical (same predicate preconditions and postconditions) definitions for \textit{all} of these ground-truth operators, except for a single operator that is defined disjunctively (the ground-truth \textit{Slice} skill specifies either of two types of knives), which we occasionally learn as two distinct operators or only recover with one of these two types. 

We also inspect the learning trajectory and find that, through the interactive learning loop, we successfully \textit{reject} many initially proposed operator definitions sampled from the language model that turn out to be redundant (which would make high-level planning inefficient), inaccurate (including apriori reasonable proposals that do not fit the environment specifications, such as proposing to \textit{clean} objects with just a \textit{towel}, when our goal verifiers require washing them with water in a \textit{sink}), or underspecified (such as those that omit key preconditions, yielding under-decomposed high-level task plans that make low-level planning difficult). 

\xhdr{Do these actions support complex planning and generalization?} \tbl{tab:minecraft-alfred-combined} shows quantitative results from \textit{n=3} randomly-initialized replications of all models, to account for random noise in sampling from the language model and stochasticity in the underlying environment (eg. in the Unity environment in ALFRED). On Minecraft, where goal specification is completely clear due to the synthetic language, we solve all tasks in each evaluation variation, including the challenging \textit{Compositional} setting --- the action libraries learned from simpler mining/crafting tasks generalize completely to complex tasks that require crafting all intermediate resources and tools from scratch. On ALFRED, we vastly outperform all other baselines, demonstrating that the learned operators are much more effective for planning and compose generalizably to more complex tasks. We qualitatively find that failures on ALFRED occur for several reasons. One is \textit{goal misspecification}, when the LLM does not successfully recover the formal goal predicate (often due to ambiguity in human language), though we find that on average, 92\% of the time, the ground truth goal appears as one of the top-4 goals translated by the LLM. We also find failures due to low-level \textit{policy inaccuracy}, when the learned policies fail to account for low-level, often geometric details of the environment (eg. the learned policies are not sufficiently precise to place a tall bottle on an appropriately tall shelf). More rarely, we see planning failures caused by slight \textit{operator overspecification} (eg. the \textit{Slice} case discussed above, in which we do not recover the specific disjunction over possible knives that can be used to slice.) Both operator and goal specification errors could be addressed in principal by sampling more (and more diverse) initial proposals. 

\xhdr{How does our approach compare to using the LLM to predict just goals, or predict task sequences?} As shown in \tbl{tab:minecraft-alfred-combined}, our approach vastly outperforms the \textbf{Low-level Planning Only} baseline on both domains, demonstrating the value of the action library for longer horizon planning. We also find a substantial improvement over the \textbf{Subgoal Prediction} baseline. While the LLM frequently predicts important high-level aspects of the task subgoal structure (as it does to propose operator definitions), it frequently struggles to robustly sequence these subgoals and predict appropriate concrete object groundings that correctly obey the initial problem conditions or changing environment state. These errors accumulate over the planning horizon, reflected in decreasing accuracy on the compositional Minecraft tasks (on ALFRED, this baseline struggles to solve any more than the basic pick-and-place tasks, as the LLM struggles to predict subgoals that accurately track whether objects are in appliances or whether the agent's single gripper is full with an existing tool.) 

\xhdr{How does our approach compare to using the LLM to learn and predict plans using imperative code libraries?}
Somewhat surprisingly, we find that the \textit{Code Policy} prediction baseline performs unevenly and often very poorly on our benchmarks. (We include additional results in \ref{sec:appendix-llm-prompting} showing that our model also dramatically outperforms this baseline using GPT-4 as the base LLM.) We find several key reasons for the poor performance of this baseline relative to our model, each which validate the key conceptual contributions of our approach. First, the baseline relies on the LLM as the planner -- as the skills are written as general Python functions, rather than any planner-specific representation, we do not use an optimized planner like FastDownward. As with \textit{Subgoal Prediction}, we find that the LLM is not a consistent or accurate planner. While it retrieves generally relevant skills from the library for each task, it often struggles to sequence them accurately or predict appropriate arguments given the initial problem state. Second, we find that imperative code is less suited in general as a hierarchical planning representation for these domains than the high-level PDDL and low-level local policy search representation we use in our model. This is because it uses control flow to account for environment details that would otherwise be handled by local search relative to a high-level PDDL action. 
Finally, our model specifically frames the library learning objective around learning a compact library of skills that enables efficient planning, whereas our Voyager re-implementation (as in ~\cite{wang2023voyager}) simply grows a library of skills which are individually executable and can be used to solve individual, shorter tasks. Empirically, as with the original model in ~\cite{wang2023voyager}, this baseline learns \textit{hundreds} of distinct code definitions on these datasets, which makes it harder to accurately plan and generalize to more complex tasks. 
Taken together, these challenges support our overarching library learning objective and the need to learn action representations suited for structured hierarchical planning.

%% file: fig-text/minecraft-alfred-combined.tex
\begin{figure}[tp]
    \centering\small
    \includegraphics[width=\textwidth]{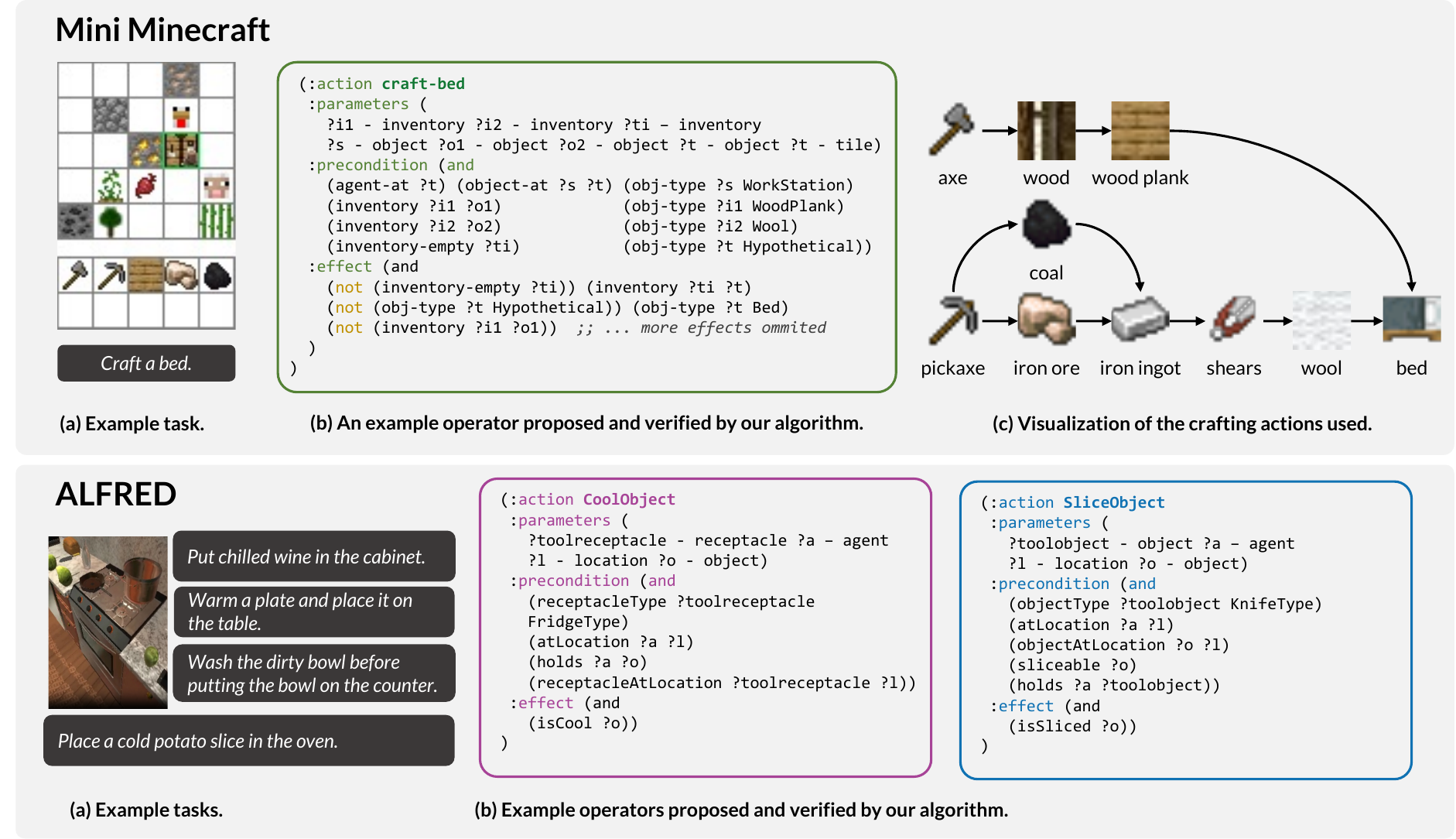}
    \vspace{-1em}
    \caption{\textit{Top}: (a) Visualization of the Mini Minecraft environment, showing an intermediate step towards \textit{crafting a bed}. (b) Operator proposed by an LLM and verified by our algorithm through planning and execution. (c) Visualization of the low-level crafting tools involved in crafting the bed. \textit{Bottom}:  (a) Visualization of the ALFRED household environment. (b) Example operators proposed by LLM and verified by our algorithm, which are composed to solve the \textit{cold potato slice} task.} \vspace{-1.2em}
    \label{fig:minecraft-alfred-combined}
    \vspace{-0em}
\end{figure}

%% file: fig-text/tab-minecraft-alfred-combined.tex
\begin{table}[tp]

\centering
\small
\resizebox{\textwidth}{!}{%
\begin{tabular}{llcccc}
\toprule
\textit{\textbf{Mini Minecraft} \textit{(n=3)}} & \textbf{LLM Predicts?} & \textbf{Library?} & \textbf{Mining}   & \textbf{Crafting}         & \textbf{Compositional}        \\ \midrule
Low-level Planning Only          & Goal                   &       \xmark          & 31\% ($\sigma$=0.0\%)              & 9\% ($\sigma$=0.0\%)              & 9\% ($\sigma$=0.0\%)   \\
Subgoal Prediction               & Sub-goals              &        \xmark           & 33\% ($\sigma$=1.6\%)             & 36\% ($\sigma$=5.6\%)                    & 6\%  ($\sigma$=1.7\%)                          \\
Code Policy Prediction           & Sub-policies           &        \cmark           & 15\% ($\sigma$=1.2\%)             & 39\% ($\sigma$=3.2\%)                     & 10\% ($\sigma$=1.7\%)                           \\
\model (Ours)                    & Goal+Operators         &         \cmark          & 100\% ($\sigma$=0.0\%)            & 100\% ($\sigma$=7.5\%)                    & 100\% ($\sigma$=4.1\%)                                       \\ \midrule
\textit{\textbf{ALFRED} \textit{(n=3 replications)}}         & \textbf{LLM Predicts?} & \textbf{Library?} & \multicolumn{3}{c}{\textbf{Original (Simple + Compositional Tasks)}}\\ \midrule
Low-level Planning Only          & Goal                   &      \xmark             & \multicolumn{3}{c}{21\% ($\sigma$=1.0\%)}                                   \\
Subgoal Prediction               & Sub-goal               &        \xmark            & \multicolumn{3}{c}{2\% ($\sigma$=0.4\%)}   \\
Code Policy Prediction           & Sub-policies           &        \cmark           & \multicolumn{3}{c}{2\% ($\sigma$=0.9\%)}     \\
\model (Ours)                    & Goal+Operators         &         \cmark          & \multicolumn{3}{c}{79\% ($\sigma$=0.9\%)}      \\ \bottomrule
\end{tabular}
}
\vspace{-1em}
\caption{(\textit{Top}) Results on \textit{Mini Minecraft}. Our algorithm successfully recovers all intermediate operators for mining and crafting, which enable generalization to more compositional tasks (which use up to 26 operators) without any additional learning. (\textit{Bottom}) Results on ALFRED. Our algorithm recovers all required household operators, which generalize to more complex compositional tasks. All results report mean performance and STD from \textit{n=3} random replications for all models.}
\label{tab:minecraft-alfred-combined}
\vspace{-1.5em}
\end{table}

%% file: text/2-related.tex
\vspace{-0.5em}
\section{Related Work} 
\vspace{-0.5em}
\xhdr{Planning for language goals.} A large body of recent work attempts to use LLMs to solve planning tasks specified in language. One approach is to directly predict action sequences~\citep{huang2022language,valmeekam2022large,silver2022pddl,wang2023describe}, but this has yielded mixed results as LLMs can struggle to generalize or produce correct plans as problems grow more complex. To combat this, one line of work has explored structured and iterative prompting regimes (\eg, `chain-of-thought` and feedback) ~\citep{mu2023embodiedgpt,silver2023generalized,zhu2023ghost}. Increasingly, other neuro-symbolic work uses LLMs to predict formal goal or action representations that can be verified or solved with symbolic planners~\citep{song2022llm,ahn2022can,xie2023translating,arora2023learning}. These approaches leverage the benefits of a known planning domain model. Our goal in this paper is to leverage language models to \textit{learn} this domain model. Another line of research aims at using LLMs to generate formal planning domain models for specific problems ~\citep{liu2023llmp} and subsequently uses classical planners to solve the task. However, they are not considering generating grounded or hierarchical actions in an environment and not learning a library of operators that can be reused across different tasks. More broadly, we share the broad goal of building agents that can understand language and execute actions to achieve goals~\citep{tellex2011understanding,misra2017mapping,nair2022learning}. See also \citet{luketina2019survey} and \citet{tellex2020robots}.

\xhdr{Learning planning domain and action representations from language.} Another group of work has been focusing on learning latent action representations from language~\citep{corona-etal-2021-modular,andreas2017modular,jiang2019language,sharma2021skill,luo2023learning}. Our work differs from them in that we are learning a planning-compatible action abstraction from LLMs, instead of relying on human demonstrations and annotated step-by-step instructions. The more recent \citet{wang2023voyager} adopts a similar overall problem specification to ours, and learns libraries of actions defined as imperative code-based policies. Our results show that learning planning abstractions enables better integration with hierarchical planning, and, as a result, better performance and generalization to more complex problems.  Other recent work ~\citep{nottingham2023embodied} seeks to learn an environment world model from interactive experience, represented as a task dependency graph; we seek to learn a richer state transition model (which represents the effects of actions) decomposed as operators that can be formally composed to verifiably satisfy arbitrarily complex new goals.

\xhdr{Language and code.} In addition to ~\cite{wang2023voyager}, a growing body of work in program synthesis, both by learning lifted program abstractions that compress longer existing or synthesized programs \citep{bowers2023top,ellis2023dreamcoder,wong2021leveraging,cao2023babble}. These approaches (including ~\cite{wang2023voyager}) generally learn libraries defined over imperative and functional programming languages, such as LISP and Python. Our work is closely inspired by these and seeks to learn representations suited specifically to solving long-range planning problems.

\xhdr{Hierarchical planning abstractions.} The hierarchical planning knowledge that we learn from LLMs and interactions in the environments are related to hierarchical task networks~\citep{erol1994htn,nejati2006learning}, hierarchical goal networks~\citep{alford2016hierarchical}, abstract PDDL domains~\citep{konidaris2018skills,bonet2019learning,chitnis2021glib,asai2020learning,Mao2022PDSketch}, and domain control knowledge~\citep{de2011learning}. Most of these approaches require manually specified hierarchical planning abstractions; others learn them from demonstrations or interactions. By contrast, we leverage human language to guide the learning of such abstractions.

%% file: text/5-conclusion.tex
\vspace{-0.5em}
\section{Discussion and Future Work}
\vspace{-0.5em}
Our evaluations suggest a powerful role for language within AI systems that form complex, long-horizon plans --- as a deep source of background knowledge about the right \textit{action abstractions} for everyday planning domains, which contains broad human priors about how the world works, how to decompose tasks, and what tasks an agent may want to solve in the future. A core goal of this paper was to demonstrate how we might integrate knowledge in language alongside the powerful search, grounding, and verification toolkits developed in hierarchical planning.

We leave open many possible extensions towards future work. Key \textbf{limitations} of our current framework point towards important directions for further integrating LMs and hierarchical planning to scale our approach: here, we build on an existing set of pre-defined symbolic predicates for initially representing the environment state; do not yet tackle fine-grained, geometric motor planning; and use a general LLM (rather than one fine-tuned for extended planning). \textbf{Future work} might generally tackle these problems by further asking how else linguistic knowledge and increasingly powerful or multimodal LLMs could be integrated here: to \textit{propose} useful named predicates over initial perceptual inputs (\eg, images) \citep{migimatsu2022grounding}; or to speed planning by bootstrapping hierarchical planning abstractions using the approach here, but then to progressively transfer planning to another model, including an LLM, to later compose and use the learned representations.

%% file: fig-text/gpt4-tab-minecraft-alfred-combined.tex
\begin{table}[tp]

\centering
\small
\resizebox{\textwidth}{!}{%
\begin{tabular}{llcccc}
\toprule
\textit{\textbf{Mini Minecraft}} & \textbf{LLM Predicts?} & \textbf{Library?} & \textbf{Mining}   & \textbf{Crafting}         & \textbf{Compositional}        \\ \midrule
Code Policy Prediction           & Sub-policies           &        \cmark           & 12\%              & 37\%                      & 11\%                            \\
Ours                             & Goal+Operators         &         \cmark          & 100\%             & 100\%                     & 100\%                                        \\ \midrule
\textit{\textbf{ALFRED} \textit{(n=3 replications)}}         & \textbf{LLM Predicts?} & \textbf{Library?} & \multicolumn{3}{c}{\textbf{Original (Simple + Compositional Tasks)}}\\ \midrule
Code Policy Prediction           & Sub-policies           &        \cmark           & \multicolumn{3}{c}{11\%}     \\
Ours                             & Goal+Operators         &         \cmark          & \multicolumn{3}{c}{70\%}      \\ \bottomrule
\end{tabular}
}
\vspace{-1em}
\caption{\textbf{Results with GPT-4 as the LLM backbone}: On both \textit{Mini Minecraft} (\textit{Top}) and ALFRED (\textit{Bottom}), our algorithm recovers all required operators, which generalize to more complex compositional tasks. Switching to GPT-4 does not impact performance trends observed across the \emph{Code as Policies (Voyager)} baseline and our method.}
\label{tab:minecraft-alfred-combined}
\vspace{-1.5em}
\end{table}